\documentclass[lettersize,journal]{IEEEtran}
\usepackage{amsmath,amsfonts}
\usepackage{algorithmic}
\usepackage{algorithm}
\usepackage{array}
\usepackage{soul}
\usepackage{xcolor}
\usepackage{color}
\usepackage[caption=false,font=normalsize,labelfont=sf,textfont=sf]{subfig}
\usepackage{textcomp}
\usepackage{stfloats}
\usepackage{url}
\usepackage{csquotes}
\usepackage{multirow}
\usepackage{booktabs}
\usepackage{diagbox}
\usepackage{verbatim}
\usepackage{graphicx}
\usepackage{cite}
\sethlcolor{yellow}

\soulregister{\cite}{7}
\soulregister{\ref}{7}
\soulregister{\pageref}{7}
\begin{document}

\title{HARMamba: Efficient and Lightweight Wearable Sensor Human
Activity Recognition Based on Bidirectional Mamba}

\author{Shuangjian Li, Tao Zhu,~\IEEEmembership{Member,~IEEE},
	Furong Duan, Liming Chen  ~\IEEEmembership{Senior Member,~IEEE}, Huansheng Ning,~\IEEEmembership{Senior Member,~IEEE,} Christopher Nugent and Yaping Wan

%  \begin{itemize}
%     \item \textcolor{blue}{Changes for Reviewer 1 highlighted in blue}
%     \item \textcolor{red}{Changes for Reviewer 2 highlighted in red}
% \end{itemize}
        % <-this % stops a space
\thanks{This work was supported in part by the National Natural Science Foundation of China (62006110), the Natural Science Foundation of Hunan Province
(2024JJ7428, 2023JJ30518) and  the Scientific research project of Hunan Provincial Department of Education (22C0229) . (Corresponding author: Tao Zhu.)}% <-this % stops a space
\thanks{Shuangjian Li, Tao Zhu, Furong Duan and Yaping Wan are with the School of Computer Science, University of South China, 421001 China (e-mail: sjli@stu.usc.edu.cn, tzhu@usc.edu.cn, frduan@stu.usc.edu.cn, ypwan@aliyun.com). }

\thanks{Liming Chen was with School
of Computer Science and Technology, Dalian University of Technology,
China. Huansheng Ning was Department of Computer \& Communication
Engineering, University of Science and Technology Beijing, 100083 China. Christopher Nugent are with the School of Computing, Ulster University, Jordanstown, Northern Ireland, UK.
(email: limingchen0922@dlut.edu.cn, ninghuansheng@ustb.edu.cn, cd.nugent@ulster.ac.uk).}}

\markboth{Journal of \LaTeX\ Class Files,~Vol.~14, No.~8, August~2021}%
{Shell \MakeLowercase{\textit{et al.}}: A Sample Article Using IEEEtran.cls for IEEE Journals}


\maketitle
% Wearable sensor human activity recognition (HAR) is a crucial area of research in activity sensing. While transformer-based temporal deep learning models have been extensively studied and implemented, their large number of parameters present significant challenges in terms of system computing load and memory usage, rendering them unsuitable for real-time mobile activity recognition applications.
% Recently, an efficient hardware-aware state space model (SSM) called Mamba has emerged as a promising alternative. Mamba demonstrates strong potential in long sequence modeling, boasts a simpler network architecture, and offers an efficient hardware-aware design. Leveraging SSM for activity recognition represents an appealing avenue for exploration.
\begin{abstract}
Wearable sensor-based human activity recognition (HAR) is a critical research domain in activity perception. However, achieving high efficiency and long sequence recognition remains a challenge. Despite the extensive investigation of temporal deep learning models, such as CNNs, RNNs, and transformers, their extensive parameters often pose significant computational and memory constraints, rendering them less suitable for resource-constrained mobile health applications. This study introduces HARMamba, an innovative light-weight and versatile HAR architecture that combines selective bidirectional State Spaces Model and hardware-aware design. To optimize real-time resource consumption in practical scenarios, HARMamba employs linear recursive mechanisms and parameter discretization, allowing it to selectively focus on relevant input sequences while efficiently fusing scan and recompute operations. The model employs independent channels to process sensor data streams, dividing each channel into patches and appending classification tokens to the end of the sequence. It utilizes position embedding to represent the sequence order. The patch sequence is subsequently processed by HARMamba Block, and the classification head finally outputs the activity category. The HARMamba Block serves as the fundamental component of the HARMamba architecture, enabling the effective capture of more discriminative activity sequence features. HARMamba outperforms contemporary state-of-the-art frameworks, delivering comparable or better accuracy with significantly reducing computational and memory demands. It's effectiveness has been extensively validated on 4 publically available datasets namely PAMAP2, WISDM, UNIMIB SHAR and UCI. The F1 scores of HARMamba on the four datasets are 99.74\%, 99.20\%,   88.23\% and 97.01\%, respectively.
\end{abstract}

\begin{IEEEkeywords}
Human Activity Recognition, Selective State Space Models, Wearable Sensors, Deep Learning, Light-weight
\end{IEEEkeywords}

\section{Introduction}
\IEEEPARstart{R}{eal-time} human activity recognition using wearable sensors has emerged as a prominent research focus in recent years\cite{gu2021survey, wang2023negative}, with extensive applications in healthcare, assisted living, and smart homes. While traditional machine learning methods, such as support vector machines (SVM), k-nearest neighbor (KNN), and random forests, have demonstrated promising results, their reliance on manually crafted features and domain expertise often limits their performance. As computing power advances, deep learning, specifically convolutional neural networks (CNNs)\cite{xu2021human}, long short-term memory (LSTM) models \cite{zhang2020data}, and Transformers \cite{dirgova2022wearable}, has revolutionized human activity recognition. However, these three models also have their own shortcomings in sequence modeling. The fixed convolution kernel size of CNNs may impede their ability to capture long-distance dependencies when processing long sequences. Additionally, the sheer number of network parameters in deep convolutional networks may restrict their applicability in resource-limited environments or on small devices. LSTMs are designed to address the long-term dependency issue, yet, in certain instances, they may still encounter challenges in effectively capturing longer-distance dependencies, potentially leading to vanishing or exploding gradient problems. Furthermore, the LSTM model contains more computational steps than the CNNs model in long sequence tasks, resulting in increased training and inference calculations. Transformers, renowned for their success in  natural language processing (NLP)\cite{waswani2017attention}, computer vision (CV)\cite{dosovitskiy2020vit}, and time series analysis\cite{ahmed2023transformers}, excels in capturing relationships and sequence correlations through the attention mechanism. However, the traditional transformer models face challenges in processing large sensor data sequences due to quadratic self-attention complexity, hindering the extraction of valuable information from individual time points. Limited training data is another issue, particularly in activity recognition where scarcity affects performance. Consequently, designing a novel architecture that balances long-term sensor signal dependencies with linear computational resource utilization is a key challenge in the field of activity recognition.

Recent research efforts\cite{gu2021efficiently, gu2021combining} have sparked significant interest in state space models (SSMs). Modern iterations of SSMs excel effectively in capturing long-range dependencies, and the incorporation of parallel training computations enables continuous enhancement over extended durations. These models exhibit a remarkable capability in modeling sequence dependencies. Various SSM-based methodologies have been introduced for applications in domains dealing with continuous signal data, notably the Linear State Space Layer (LSSL) \cite{gu2021combining}, Structured State Space Sequence Model (S4) \cite{gu2021efficiently}, Diagonal State Space (DSS) \cite{gupta2022diagonal}, and S4D \cite{gu2022parameterization}. Leveraging efficient convolution or recursion computations with linear or near-linear scalability with respect to sequence length, these approaches are proved versatile in processing sequence data across diverse tasks and modalities. Notably, the recent innovation of Mamba \cite{gu2023mamba} introduced a straightforward selection mechanism to discretizes SSM parameters based on input, along with a hardware-aware algorithm that achieves highly efficient training and inference. Nevertheless, the exploration of SSM-based backbone networks for processing IMU activity data remains an uncharted territory.

While Mamba has demonstrated success in modeling language, audio, video, and other domains, it encounters challenges when applied to sensor activity data due to its unidirectional modeling approach and lack of consideration for sequence position information. Inspired by \cite{gu2023mamba}, we introduce the HARMamba model. Initially, we segment each channel of continuous sensor data into patches, with each patch encapsulating sensor channel information. Subsequently, we apply position embeddings to each patch block, projecting them as linear vectors into Mamba blocks. Since Mamba's one-way modeling lacks contextual awareness, these Mamba blocks leverage bidirectional selective state space to efficiently compress patch sequence representations, with position embeddings facilitating information exchange between patch blocks and enhancing the model's accuracy in activity recognition. Leveraging Mamba's linear modeling capability, HARMamba adopts a pure SSM approach, modeling IMU signals in patch form.  Our approach maintains the training parallelism of SSM, ensures inference efficiency, and achieves significant enhancements with a low number of parameters. This design establishes an efficient backbone network for activity recognition.

The main contributions of this paper can be summarized as follows:
\begin{itemize} 
	\item We introduced a simple and light-weight framework called HARMamba, which is a pioneering approach that applies the selective state space model and scanning mechanism in activity recognition tasks.
	\item The HARMamba framework adopts a sensor data channel-independent approach and patch operation to process sensor signal sequences, modeling the global context and location embedding of sensor signal patch sequences through bidirectional SSM.
	\item Our proposed framework achieves light-weight utilisation of computational resources, with higher computational speed, lower parameters and memory consumption than current state-of-the-art work. This feature facilitates its deployment on real-time activity detection devices.
        \item Performing activity recognition classification experiments on four widely utilized benchmark datasets, the findings demonstrate that HARMamba demonstrates higher recognition performance when compared to alternative frameworks.
\end{itemize}

The remainder of this paper is structured as follows: Section II reviews related work. Section III details the four datasets employed and the preprocessing methods applied. Section IV provides a comprehensive description of the adopted model architecture and its components. Section V presents experimental results evaluating the HARMamba model on four public datasets. Finally, the concluding section discusses the research findings and outlines future work.
\section{Related Work}
This section provides a review of several related works on sensor-based activity recognition and state space models.
\subsection{Sensor-based activity recognition}
Sensor-based human activity recognition primarily relies on IMU sensors in wearable devices like smartwatches, smart insoles, and exercise bands. Traditional approaches, such as fixed-size sliding windows and dense labeling, have demonstrated strong performance \cite{chen2021deep}. Recent advancements in deep learning have facilitated the application of convolutional neural networks\cite{yang2015deep}, recurrent neural networks\cite{chen2019semisupervised}, and hybrid structures \cite{ordonez2016deep} for feature extraction and activity recognition. For instance, \cite{9055403} employed an LSTM-based method to identify fine-grained patterns using high-level features from sequential motion data, \cite{9732352} introduced a deformable convolutional network for activity recognition from complex sensory data, and \cite{mim2023gru} presented a GRU-INC model that initializes attention-based GRUs to effectively leverage spatiotemporal information in time series. Additionally, \cite{challa2022multibranch} proposed a multi-branch CNN-BiLSTM network that extracts features from raw sensor data with minimal preprocessing. Inspired by image semantic segmentation, \cite{badrinarayanan2017segnet, yao2018efficient, zhang2019human} employed FCNs and U-nets for dense label prediction, achieving commendable recognition rates. \cite{zhang2021conditional} further developed Conditional-UNet, which models conditional dependencies between dense labels for coherent HAR.

The Transformer's success in sequence modeling has led to a surge in the application of attention mechanisms to sensor activity sequences, significantly enhancing activity recognition performance. \cite{khan2021attention} introduced an attention-based Multi-head model for human activity recognition, demonstrating its effectiveness. \cite{tang2022triple} further advanced the field by proposing a triplet cross-dimensional attention method, which employs three branches to capture interactions between sensor dimensions, time, and channel aspects in sensor-based tasks. \cite{gao2021deep} introduced a deep CNNs with an attention mechanism, allowing for kernel selection across multiple branches with distinct receptive fields in the context of human activity recognition. \cite{dirgova2022wearable} showcased the potential of adaptive Transformers with their model, achieving favorable results on smartphone motion sensor data from diverse activities. Lastly, \cite{al2023multi} presented the Multi-ResAtt model, integrating recurrent neural networks and attention for time series feature extraction and activity recognition.

Deep learning models have demonstrated exceptional performance in human activity recognition, but their resource-intensive nature often hinders deployment on edge devices. To address this issue, researchers have proposed light-weight deep learning models, for instance, \cite{AGARWAL20202364} proposes a HAR light-weight deep learning model that requires less computing power, making it suitable for deployment on edge devices.  \cite{10100934} employs pruning and quantization techniques to CNNs, effectively reducing computational demands and memory consumption. \cite{huang2023channel} further contributes by introducing a light-weight CNNs designed for activity recognition, achieving a balance between recognition accuracy and resource efficiency.
\subsection{State space models}
  SSMs \cite{gu2021efficiently, gu2021combining} have emerged as a promising architecture for sequence modeling, with Mamba \cite{gu2023mamba} standing out as a selective SSM that integrates time-varying parameters and employs a hardware-aware algorithm for efficient training and inference. Mamba employs selective scan algorithms and recomputation techniques optimizing the I/O and memory efficiency of modern GPU hardware accelerators. To address memory constraints, it leverages recomputation minimizing the memory footprint associated with selective state selection methods. In its forward pass, Mamba bypasses storing intermediate states, allowing for direct gradient computation and transmission. As the network propagates backward, Mamba swiftly recomputes these states, thereby minimizing the expense of High bandwidth memory(HBM) data reads, resulting in a highly efficient process. Mamba's success in long sequence modeling has sparked its application in various domains, such as U-Mamba \cite{U-Mamba}, which explores Mamba's potential in vision tasks by proposing a universal segmentation network for 2D and 3D biomedical images. Vim \cite{zhu2024vision} introduces a bidirectional SSM block for efficient visual representation learning, achieving competitive performance with ViT \cite{dosovitskiy2020vit}. VMamba \cite{liu2024vmamba} introduces the Cross Scan Module (CSM), converting visual images into ordered sequences with linear complexity and global receptive field. LightM-UNet \cite{liao2024lightm} combines Mamba and UNet in a light-weight framework for efficient image processing. 2D SSM \cite{baron20232} integrates SSM blocks into Transformer blocks \cite{dosovitskiy2020vit, waswani2017attention}, while DenseMamba \cite{he2024densemamba} enhances SSMs by selectively fusing shallow hidden states for fine-grained information retention. ConvSSM \cite{smith2024convolutional} combines ConvLSTM \cite{shi2015convolutional} principles with SSM, enabling parallel scans and sub-quadratic parallelization. TimeMachine \cite{ahamed2024timemachine} utilizes Mamba for long-term dependencies in multivariate time series, and Motion Mamba \cite{zhang2024motion} introduces a motion generation model with hierarchical temporal Mamba (HTM) and bidirectional spatial Mamba (BSM) blocks for temporal and bidirectional processing of latent pose data.
\section{DATA DESCRIPTION}

This section provides a detailed introduction to four public datasets and the preprocessing methods employed for each. Table \ref{Specifics of the baseline data set} provides details of the four datasets.

\subsection{WISDM \cite{kwapisz2011activity}}  
The data were obtained by 29 participants using phones with triaxial acceleration sensors placed in their trouser pockets with a sampling frequency of 20 Hz. Walking, strolling, walking up stairs, walking down stairs, standing motionless, and standing up were among the six daily activities undertaken by each participant. Fill missing values in a dataset using linear interpolation.

% this dataset was collected from 29 participants who wore mobile phones equipped with three-axis accelerometers in their \st{pants} \textcolor{red}{trouser} pockets. The data were sampled at 20 Hz. The study participants engaged in six activities daily: walking, strolling, ascending stairs, descending stairs, standing, and standing still. To address any missing data in the dataset, the corresponding column's mean is used for filling.

\subsection{PAMAP2 \cite{reiss2012introducing}}  This dataset comprises data collected from nine participants wearing IMUs on their chest, hands, and ankles. These IMUs gather information on acceleration, angular velocity, and magnetic sensors. Each participant completed 12 mandatory activities, including lying down, standing, and going up and down stairs, as well as 6 optional activities, such as watching TV, driving a car, and playing ball.

\subsection{UNIMIB SHAR\cite{micucci2017unimib}}  The dataset was collected from the University of Milano-Bicocca. Thirty volunteers, aged 18 to 60, had their Samsung phones equipped with Bosch BMA 220 sensors. The study recorded 11,771 activities by placing phones in the participants' left and right pockets. The dataset comprises 17 categories of activities, divided into two groups: 9 activities of daily living (ADL) and 8 falling behaviors. Each activity was repeated 3 to 6 times.

\subsection{UCI \cite{anguita2013public}}  The dataset were generated by 30 participants aged 19 to 48 years performing daily activities. Each participant wore a Samsung Galaxy S2 smartphone around their waist. They collected sensor data in 9 dimensions using the phone’s built-in acceleration, gyroscope, linear acceleration, and 3-axis angular velocity sensors. Six different daily activities (walking, walking upstairs, walking downstairs, standing still, standing, and lying down) were performed.

\begin{table}[htbp]
	\renewcommand\arraystretch{1.5}
	\centering 
	\caption{\label{Specifics of the baseline data set}Specifics of the baseline dataset}
	\begin{tabular}{lccccc}
		\toprule
		 Dataset & Categories & Subject & Rate & Window & Sample \\
		\midrule
    WISDM&  6 & 29 & 20HZ & 200 & 1,098,208\\
		PAMAP2 & 12 & 9 & 33.3HZ & 512 & 2,872,533 \\
		UNIMIB SHAR & 17 & 4 & 30HZ & 151 & 11,771\\
		UCI & 6 & 20 & 50HZ & 128 & 74,8406\\
		\bottomrule
	\end{tabular}
\end{table}

We applied standard preprocessing techniques, including the imputation of missing values using linear interpolation , standardization, and the segmentation of the data set using a 50\% overlapping sliding window, to preserve the temporal coherence between data points. For the partitioning of the data set, we employed the mean method to divide the continuous data set into training, validation, and test sets, in the proportions of 0.7, 0.1, and 0.2, respectively.
\section{METHOD}

This section provides a detailed introduction to the HARMamba architecture. Firstly, the basic concepts of SSM are presented, followed by an explanation of how the input token sequence is processed and the HARMamba model is utilized for HAR.

\begin{figure*}[htb]
	\centering
	\includegraphics[width=\textwidth]{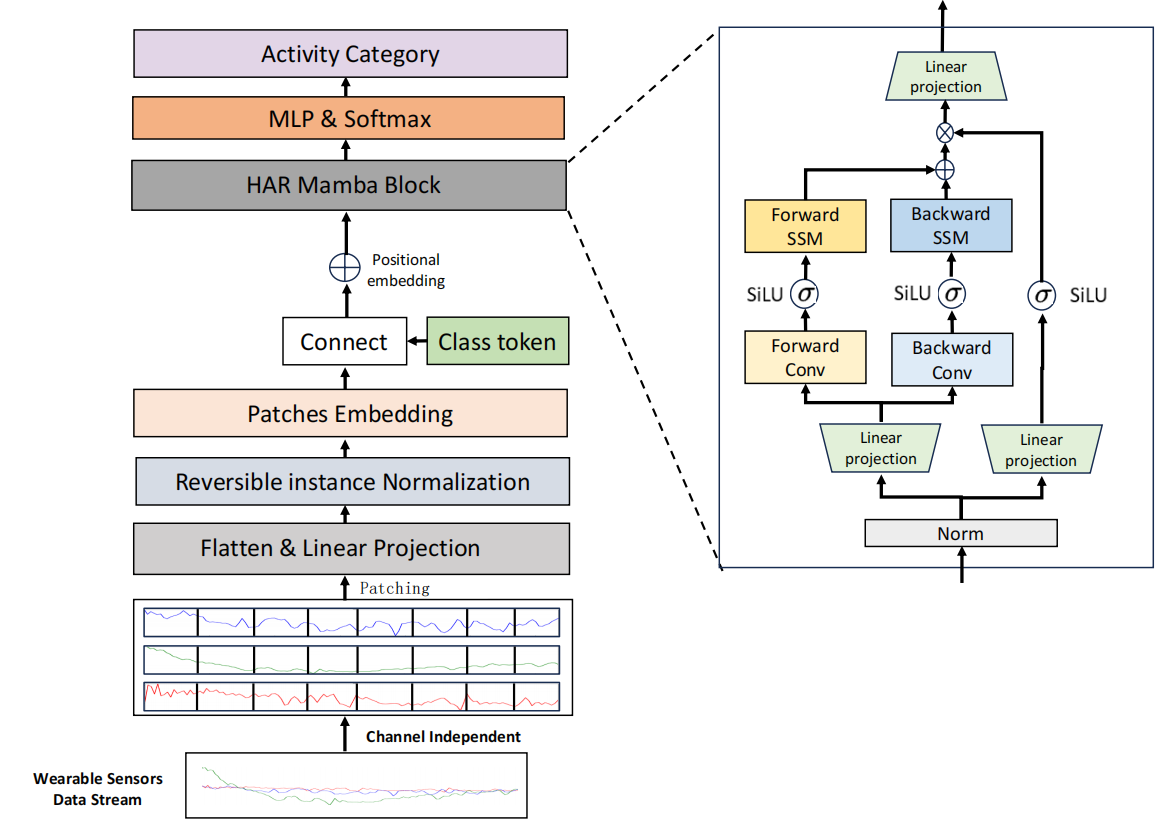}
	\caption{illustrates the schematic of our proposed HARMamba model. Initially, we render the input sensor signal sequence channel-agnostic, segment the sequence of each channel into patch sequences, and subsequently map them to patch tokens while incorporating position encoding for each token. Subsequently, the token sequence is fed into the HARMamba architecture. In contrast to Mamba, which is tailored for text sequences, the HARMamba encoder executes bidirectional processing of token sequences.}
	\label{recongnitionmodel}
\end{figure*}

\subsection{Preliminaries}

Inspired by the Structured State Space Sequence Model (S4) \cite{gu2021efficiently} and Mamba \cite{gu2023mamba}, SSM has emerged as a promising architecture for efficient long sequence modeling. Structured SSM maps a sequence $x(t)\in \mathbb{R}^{L}$ to $y(t)\in \mathbb{R}^{L}$ by hiding the state $h(t)\in \mathbb{R}^{N}$, and the model uses $\textbf{A} \in \mathbb{R}^{N\times N}$ as an evolution parameter and $\textbf{B} \in \mathbb{R}^{N\times 1}$ and $\textbf{C} \in \mathbb{R}^{1\times N}$ as projection parameters. The whole process can be formulated as:

\begin{equation}
\begin{aligned}
h^{'}(t) &= \textbf{A}h(t) + \textbf{B}h(t)  \\
y(t) &=\textbf{C}h(t)\label{eq1}
\end{aligned}
\end{equation}

To apply SSM to discrete data, the continuous parameters $\textbf{A}$ and $\textbf{B}$ are converted into discrete parameters $\overline{\textbf{A}}$ and $\overline{\textbf{B}}$ using the time scale parameter $\Delta$. One commonly used transformation method is the zero-order hold (ZOH), which is defined as follows:

\begin{equation}
\begin{aligned}
\overline{\textbf{A} } &= exp(\Delta\textbf{A} ) \\
\overline{\textbf{B} } = (\Delta\textbf{A} )&^{-1}(exp(\Delta\textbf{A} )-I)\cdot \Delta\textbf{B} \label{eq2}
\end{aligned}
\end{equation}

After discretizing the parameters, the Eq. (\ref{eq1}) can be rewritten using the discretization parameters as follows:

\begin{equation}
\begin{aligned}
h_{t} &= \overline{ \textbf{A}}h_{t-1} + \overline{ \textbf{B}}x_{t}  \\
y_{t} &=\textbf{C}h_{t} \label{eq3}
\end{aligned}
\end{equation}

The model calculates the output using global convolution:
\begin{equation}
\begin{aligned}
\overline{\textbf{K} } = (\textbf{C}\overline{\textbf{B} }&, \textbf{C}\overline{\textbf{AB} } ,\dots,\textbf{C}\overline{\textbf{A} }^{L-1}\overline{\textbf{B} } ) \\
y&=\textbf{x}*\overline{\textbf{K}}\label{eq4}
\end{aligned}
\end{equation}

The length of the input sequence $\mathbf{x}$ is denoted by $L$, and the structured convolution kernel is represented by $\overline{\textbf{K}} \in \mathbb{R}^{L}$. Eq. (\ref{eq3}) is utilized for autoregressive inference of the model, while Eq. (\ref{eq4}) is effective for parallel training.

\emph{Discretization}: Modern deep learning frameworks and hardware are typically designed for discrete-time operations. Therefore, after discretizing the SSM, it can be transformed into a model that runs efficiently on these platforms. Specifically, we employ the ZOH method for discretization operations to convert continuous parameters into their discretized counterparts, as illustrated in eq. (\ref{eq2}). In this context, $\Delta$ represents a critical parameter of the selection mechanism output, which is utilized to adjust the step size of the state update. This adjustment influences the extent of state change in the system at each time step, as well as the calculation of the state transition matrix $\textbf{A}$ and the input matrix $\textbf{B}$. $\Delta$ also plays a role in achieving selective propagation and forgetting of information, a larger $\Delta$ value indicates that the system responds more strongly to the current input, implying a selective retention of information regarding the current input. Conversely, a smaller $\Delta$ value suggests that the system responds less to the current input and relies more on the previous state $h_{t-1}$.

\emph{Linear recursive mechanism} :Selective SSM employs a recursive structure, where the state at each time step, $h_t$, is a combination of the state from the previous time step, $h_{t-1}$, and the current input. The precise formula is shown in eq. (\ref{eq3}). This recursive characteristic guarantees that the current state not only relies on the current input but also incorporates historical information, enabling the model to capture the contextual information of the input at various positions.

\subsection{HAR Mamba}

Figure \ref{recongnitionmodel} shows the definition of the $D_c$-dimensional input wearable sensor sequence as $\textbf{X}_{1:L} = \left[x_1,\dots,x_{L}\right],x_i \in \mathbb{R}^{D_c}$. Here, $x_i$ represents the sensor signal collected at timestamp t, and the activity label corresponding to each sample is marked as $y_{1:L}=\left[ y_1,\dots,y_L\right]$, $y_t \in \mathbb{R}^C$. $C$ represents the category of the activity number, $t$ is the time step, $L$ is the length of the wearable sensor sequence, and the length of Patch is defined as $P$.

The input sensor sequence $\left[x_1, \dots, x_L \right]$ is first divided into $D_c$ single-variable sequences $x^{(i)}\in \mathbb{R}^{1\times L}$. Prior to processing the patches, we apply reversible instance normalization (RevIN)\cite{kim2021reversible} to multiple channels of sensor signal data. This is intended to address the issue of non-uniform time distribution between training and test data, which is often referred to as distribution shift. Before combining the channel modules, we normalize the data for each channel using instance normalization. For each instance of $x^{(i)}$, we compute the mean and standard deviation. 

\begin{equation}
\begin{aligned}
RevIN(x^{(i)})=\left \{  \gamma_i \frac{x^{(i)}-Mean(x^{(i)})}{\sqrt{Var(x^{(i)})+\varepsilon } } \right \},i=1,2,\dots,D_c   \label{eq5}
\end{aligned}
\end{equation}

After normalization of reversible instances, each single-channel sensor signal sequence $x^{(i)}$ can be further divided into overlapping or non-overlapping Patch blocks, with a length of P. The non-overlapping part between two consecutive Patch blocks is referred to as the step size, expressed as S. The segmentation of the patch block results in a patch-level sensor sequence $x_{p}^{(i)} \in \mathbb{R}^{P \times N}$, where $N=\left \lfloor  \frac{(L-P)}{S}\right \rfloor $. In order to facilitate the classification of sensor sequences, a class token ($token_{cls}$) has been introduced, which performs global feature aggregation over all other tokens. The $token_{cls}$ is initially set to an all-zero matrix and is updated continuously as the network trains. The input token and $token_{cls}$ are spliced together to create a patch token vector of length $N+1$. Subsequently, we linearly project this sequence into a D-dimensional vector space. Each token is then augmented with a position embedding $E_{pos}\in \mathbb{R}^{(N+1)\times D}$. This process is expressed mathematically as:

\begin{equation}
\begin{aligned}
\textit{Tokens} = \textit{Concat}(x_p^{(i)}\textbf{W},token_{cls}), i=1,2,\dots,N \label{eq6}
\end{aligned}
\end{equation}

\begin{equation}
\begin{aligned}
\textbf{T}_0 = \textit{Tokens} + E_{pos} \label{eq7}
\end{aligned}
\end{equation}

Where $\textbf{W}\in \mathbb{R}^{(N\times L)\times D}$ is the learnable projection matrix, then we input the token sequence representation $\textbf{T}_{l-1}$ into the HARMamba Block of the $l$-th layer, and the output is $\textbf{T}_l$. We normalize the output class token $\textbf{T}_{l}^{N+1}$ and then transport it to the MLP. Last, the activity class prediction $\hat{y}$ is obtained through the Softmax layer. The specific representation is as follows:

\begin{equation}
\begin{aligned}
\textbf{T}_l = \textit{HARMambaBlock}(\textbf{T}_{l-1}) \label{eq8}
\end{aligned}
\end{equation}

\begin{equation}
\begin{aligned}
\hat{y} = Softmax(MLP(Norm(\textbf{T}_l^{N+1}))) \label{eq9}
\end{aligned}
\end{equation}

Predicted and true labels are used to construct an activity classification loss, which efficiently optimizes model parameters. The loss is defined as follows:

\begin{equation}
    L_{cls}(\hat{y},y ) = -\frac{1}{L} \sum_{l=1}^{L}\sum_{c=1}^{C}ylog(\hat{y} )  
\end{equation}
\subsection{HARMamba Block}
This section introduces the HARMamba Block, which models sensor sequences bidirectionally. The HARMamba Block is depicted in Figure \ref{recongnitionmodel}.

The token sequence representation $\textbf{T}_{l-1}$ is normalized using a normalization layer. The output sequence is then projected into $\textbf{x}$ and $\textbf{z}$ of size $E$ through two linear mappings.

\begin{equation}
\begin{aligned}
\textbf{x},\textbf{z} = \textit{Linear}(\textit{Norm}(\textbf{T}_{l-1})) \label{eq10}
\end{aligned}
\end{equation}

The input $\textbf{x}$ is processed through two independent branches, one forward and one backward. In each direction, $\textbf{x}$ undergoes one-dimensional causal convolution resulting in $\textbf{x}_{o}^{'}$. The SiLU activation function is then applied to $\textbf{x}_{o}^{'}$ before it is fed into the SSM. Within the SSM, $\textbf{x}_{o}^{'}$ is linearly projected onto $\textbf{B}_o$, $\textbf{C}_o$, and $\Delta_o$, and then converted through $\Delta_o$ to obtain $\overline{\textbf{A}}_o$ and $\overline{\textbf{B}}_o$. Finally, $y_{forward}$ and $y_{backward}$ are obtained through bidirectional SSM.

\begin{equation}
\begin{aligned}
y_{forward/backward} = SSM(SiLU(Conv1D(\textbf{x}))) \label{eq11}
\end{aligned}
\end{equation}

Finally, we add $y_{forward}$ and $y_{backward}$, calculate the result with $\textbf{z}$ to get the gate value, and then use it through a linear projection to obtain the final output Token sequence $\textbf{T}_l$.

\begin{equation}
\begin{aligned}
\textbf{T}_l = Linear((y_{forward} + y_{backward})\otimes \textbf{z})\label{eq12}
\end{aligned}
\end{equation}

Algorithm 1 describes the specific process. $l$ represents the number of HARMamba Blocks, \emph{D} represents the hidden state dimension, and \emph{E} represents the extended state dimension.

\begin{algorithm}[htb]
	\caption{HARMamba Block}\label{Prediction}
	\renewcommand{\algorithmicrequire}{\textbf{Input:}}
	\renewcommand{\algorithmicensure}{\textbf{Output:}}
	\begin{algorithmic}[1]
		\REQUIRE   
		Patch token sequence  $\mathbf{T}_{l-1}$ : $(B,L,D)$\\ 
		\ENSURE
		Patch token sequence $\mathbf{T}_{l}$ : $(B,L,D)$\\ 
		\STATE $\mathbf{T}_{l-1}^{'}$ : $(B,L,D)$ $\leftarrow$ $\textbf{Norm($\textbf{T}_{l-1} $)}$ \\	
		\STATE $\mathbf{x,z}$ : $(B,L,E)$ $\leftarrow$ $\textbf{Linear($\textbf{T}_{l-1}^{'} $)}$
		\FOR{$ \mathbf{o}$ in \{\textit{forward, backward}\} }
		\STATE $\mathbf{x}_{o}^{'}$ : $(B,L,E)$ $\leftarrow$ $\textbf{SiLU(Conv1D(\bf{x}))}$
		\STATE $\mathbf{B}_{o},\mathbf{C}_{o},\mathbf{\Delta }_{o}$ $\leftarrow$ $\mathbf{Linear(\bf{x}_{o}^{'})}$
		\STATE $\overline{\bf{A}}_{o},\overline{\bf{B}}_{o}$ : $Transform$ $\bf{A}$,$\bf{B}$ $through$ $\bf{\Delta}_{o}$
		\STATE $\bf{y}_{o}$ : $(B,L,E)$ $\leftarrow$ $\bf{SSM(\overline{\bf{A}}_{o},\overline{\bf{B}}_{o},\bf{C}_{o})}$ 
		\ENDFOR
		\STATE $\bf{T}_{l}$ $\leftarrow$ $\bf{Linear((y_{forward} + y_{backward}) \odot \bf{SiLU(\bf{z})})}$
		\STATE Return: $\bf{T}_{l}$
	\end{algorithmic}
\end{algorithm}

\subsection{Channel independently}

HARMamba processes the data from each sensor channel independently, rather than combining the data from all channels for model training. Specifically, prior to inputting the data into the HARMamba model, we concatenate the data from all channels in channel order. During the training phase, we extract data from different channels for training purposes, and subsequently, we merge the learned feature results, which effectively treats different dimensions as independent while sharing embeddings and weights within each dimension. In comparison to channel mixture models, channel-independent models are less susceptible to overfitting. By maintaining channel independence, the model is able to concentrate on the relevant signals within each channel, thereby minimizing the influence of noise from other channels.
\section{Experiment}

This section introduces the experimental settings, datasets, recognition metrics, and competing algorithms used in the experiments. Subsequently, the proposed HARMamba framework is evaluated in terms of activity recognition accuracy and F1 score. The results demonstrate a significant advantage in recognition performance.

\subsection{Experimental setup}

The method was implemented using PyTorch and trained on two RTX 3090 GPUs. The batch size was set to 64, and AdamW with a momentum of 0.9 and a weight decay of 0.05 was used to optimize the model, along with a learning rate of 0.0001. The HARMamba block consists of 12 layers. The patch sizes for PAMAP2, UCI, UNIMIB SHAR, and WISDM are 16, 8, 10, and 4 respectively.

% \subsection{Datasets}

\subsection{Evaluation metrics}
To assess the activity categories of the final output of the model, we employ the following identification metrics:

\textbf{Accuracy}: an overall measure of correctness across all categories, calculated as follows:

\begin{equation}
    Accuracy=\frac{\sum_{c=1}^{C}TP_{c}+\sum_{c=1}^{C}TN_{c}}{\sum_{c=1}^{C}(TP_{c}+TN_{c}+FP_{c}+FN_{c})}
\end{equation}

\textbf{F1 score:} The F1 score is calculated based on the proportion of samples:

\begin{equation}
    F1=\sum_{c=1}^{C}2*w_c\frac{prec_c\cdot recall_c}{prec_c+recall_c}
\end{equation}

Where $prec_c=\frac{TP_c}{TP_c+FP_c}$ and $recall_c=\frac{TP_c}{TP_c+FN_c}$ represent precision and recall, respectively. The weight $w_c$ is calculated as the ratio of $n_c$ to $N$, where $n_c$ is the number of Patch samples per class, and $N$ is the total number of samples.

\subsection{Comparisons with activity recognition algorithm}
\textit{1) Activity recognition result}

\begin{table*}[htbp]
	\renewcommand\arraystretch{1.8}
	\centering 
	\caption{\label{activity recognition accuracy result}ACC and F1 Performance on four public datasets}
	\resizebox{\textwidth}{!}{
		\begin{tabular}{lcccc}
			\toprule
			Dataset & Related work & Method & Accuracy & F1 score \\
			\midrule
                & Challa S K et al. (2022) \cite{challa2022multibranch}& CNN-BiLSTM  & 94.29\% & 0.9427\\
                 & Tang Y et al. (2022) \cite{tang2022triple}& Triple cross domain Attention & - & 0.9320\\
			 & Essa E et al. (2023) \cite{essa2023temporal}& Temporal-channel convolution with self-attention network &89.10\% & 0.8642\\
                 & Tang Y et al. (2022) \cite{tang2022dual}& Dual-branch interactive networks on multichannel & 92.05\% & 0.9199\\
                 & Huang W et al. (2023) \cite{huang2023channel}& Channel-Equalization-HAR  & 92.14\% & 0.9218\\
            PAMAP2 & Tang Y et al. (2022) \cite{tang2022multiscale}& HS-ResNet  & 93.75\% & 0.9429\\
                 
                 & F. Duan et al. (2023) \cite{duan2023multi}& MTHAR  & 94.50\% & 0.9480\\
                 & Tong L et al. (2022) \cite{tong2022novel}& Bi-GRU-I  & 95.42\% & 0.9545\\
                 & J Li et al. (2023) \cite{li2023multi}& Multi-resolution Fusion Convolutional Network & \underline{98.14\%} & \underline{0.9812}\\
                 & Current Works & HARMamba & \bf{99.91\%} & \bf{0.9974}\\
			\bottomrule
   & Challa S K et al. (2022) \cite{challa2022multibranch}& CNN-BiLSTM  & 96.37\% & 0.9631\\
   & Tang Y et al. (2022) \cite{tang2022triple}& Triple cross domain Attention & - & 0.9677\\
    & Huang W et al. (2023) \cite{huang2023channel}& Channel-Equalization-HAR  & \underline{97.35\%} & 0.9712\\
    UCI & Tang Y et al. (2022) \cite{tang2022multiscale}& HS-ResNet  & 97.28\% & \bf{0.9738}\\
    & F. Duan et al. (2023) \cite{duan2023multi}& MTHAR  & 96.32\% & \underline{0.9723}\\
			 & MA Khatun et al. (2022) \cite{khatun2022deep}& CNN-LSTM with Self-Attention &93.11\% & 0.8961\\
                 & Y Wang et al. (2023) \cite{wang2023novel}& Multifeature extraction framework based on attention mechanism & 96.00\% & 0.9580\\
                 & Current Works & HARMamba & \bf{97.65\%} & 0.9701\\
			\bottomrule
   & Tang Y et al. (2022) \cite{tang2022triple}& Triple cross domain Attention & - & 0.7855\\
   & MAA Al-Qaness et al. (2023) \cite{al2023multi}& Multilevel Residual Network With Attention &\underline{85.35\% }& \underline{0.8506}\\
                
            UNIMIB SHAR    & Huang W et al. (2023) \cite{huang2023channel}& Channel-Equalization-HAR  & 78.65\% & 0.7878\\
             & Tang Y et al. (2022) \cite{tang2022multiscale}& HS-ResNet  & 79.02\% & 0.7919\\
                
                & F. Duan et al. (2023) \cite{duan2023multi}& MTHAR  & 76.48\% & 0.7571\\
                 & Current Works & HARMamba & \bf{88.08\%} & \bf{0.8823}\\
			\bottomrule
               & Challa S K et al. (2022) \cite{challa2022multibranch}& CNN-BiLSTM  & 96.05\% & 0.9604\\
               & Tang Y et al. (2022) \cite{tang2022triple}& Triple cross domain Attention & - & 0.9861\\
                & Essa E et al. (2023) \cite{essa2023temporal}& Temporal-channel convolution with self-attention network &92.51\% & 0.8518\\
                 & Tang Y et al. (2022) \cite{tang2022dual}& Dual-branch interactive networks on multichannel & 98.17\% & 0.9856\\
               WISDM & Huang W et al. (2023) \cite{huang2023channel}& Channel-Equalization-HAR  & \underline{99.04\%} & \underline{0.9918}\\
                  & Tong L et al. (2022) \cite{tong2022novel}& Bi-GRU-I  & 98.25\% & 0.9712\\
                & J Li et al. (2023) \cite{li2023multi}& Multi-resolution Fusion Convolutional Network & 98.60\% & 0.9864\\
                & Y Wang et al. (2023) \cite{wang2023novel}& Multifeature extraction framework based on attention mechanism & 97.90\% & 0.9800\\

                 & Current Works & HARMamba & \bf{99.11\%} & \bf{0.9920}\\
			\bottomrule
		\end{tabular}
	}
\end{table*}
% This subsection compares the recognition performance of the proposed model with state-of-the-art competing methods on four public wearable device-based HAR datasets. Our method achieves results comparable to or better than the state-of-the-art on every public dataset. The results are presented in Table \ref{activity recognition accuracy result}. Our method achieves higher recognition accuracy than other recent works on three datasets.
This section compares the recognition performance of our proposed model with contemporary competing methods on four publicly available wearable-based HAR datasets. Our model consistently outperforms or matches state-of-the-art results, as demonstrated in Table \ref{activity recognition accuracy result}. Our method exhibits notable improvements, with increases of 1.77\%, 0.30\%, 2.73\% and 0.07\% in accuracy over recent works on the PAMAP2, UCI, UNIMIB SHAR and WISDM datasets, respectively. 

\begin{figure*}[htbp]
	\centering
	\subfloat[PAMAP2 (no training)]{\includegraphics[width=1.9in]{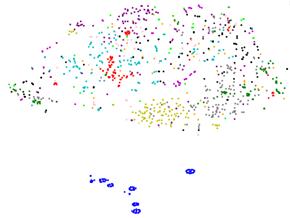}%
		\label{pamap2_none}}
	\subfloat[WISDM (no training)]{\includegraphics[width=1.9in]{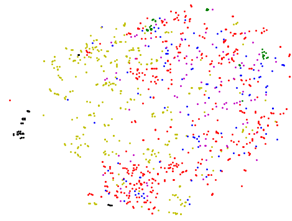}%
		\label{wisdm_none}}
	\subfloat[UNIMIB SHAR (no training)]{\includegraphics[width=1.9in]{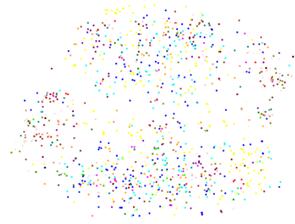}%
		\label{unimib_none}}
    \subfloat[UCI (no training)]{\includegraphics[width=1.9in]{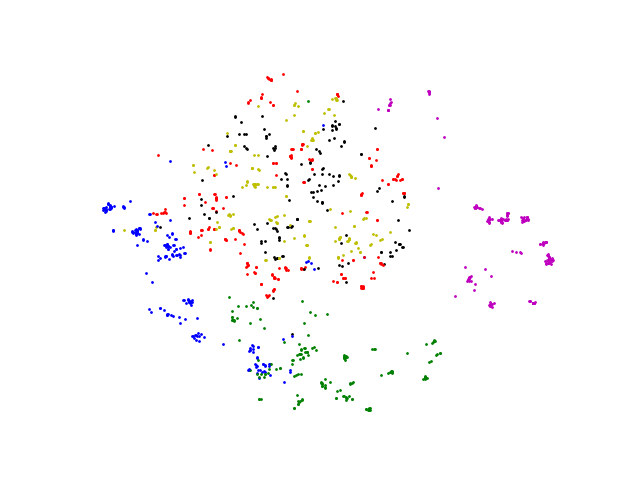}}%
		\label{uci_none}
	\hfil
	\subfloat[PAMAP2 (trained)]{\includegraphics[width=1.9in]{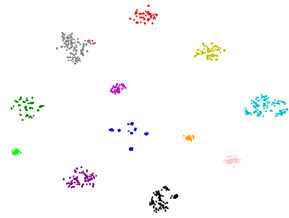}%
		\label{pamap2_trained}}
	\subfloat[WISDM (trained)]{\includegraphics[width=1.9in]{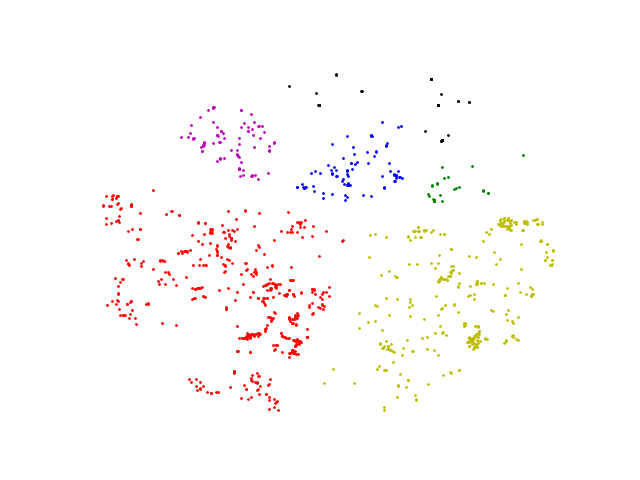}%
		\label{wisdm_trained}}
	\subfloat[UNIMIB SHAR (trained)]{\includegraphics[width=1.9in]{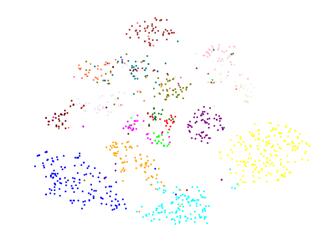}%
		\label{unimib_trained}}
  \subfloat[UCI (trained)]{\includegraphics[width=1.9in]{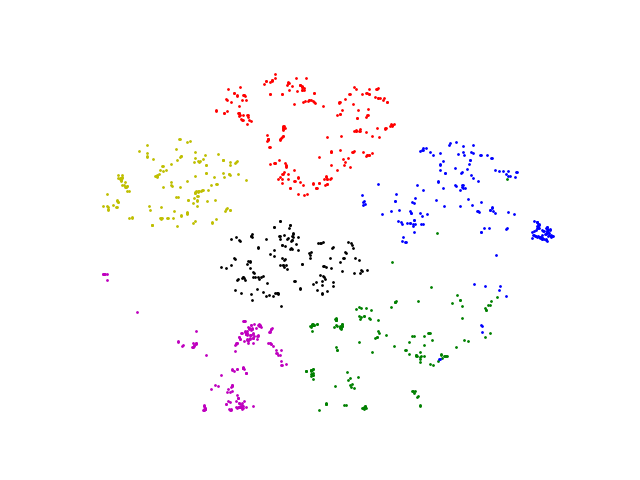}}%
		\label{uci_trained}
	\caption{t-SNE visualization results. Subfigures (a), (b), (c) and (d) display sample t-SNE visualization results without any training, while subfigures (e), (f), (g) and (h) show the t-SNE visualization results output by the pre-trained model. The t-SNE visualization results show samples from four different datasets: PAMAP2, WISDM, UNIMIB SHAR and UCI. The activity categories are represented by different colours.}
	\label{T-SNE result}
\end{figure*}

To assess the efficacy of pre-trained models in capturing and representing activity features across different datasets, we employed t-SNE visualization on HARMamba block outputs from four sources: PAMAP2, WISDM, UNIMIB SHAR and UCI. Each point in Figure \ref{T-SNE result} is color-coded according to its corresponding activity label, and the position of each label reflects the feature dimensionality reduction of the model in the 2D projection. This visualization helps distinguish different activities through assigned colors and provides a clear representation of the spatial arrangement of features. Figure \ref{T-SNE result} (a), (b), (c) and (d) show the t-SNE results of the untrained model output features; it is evident that feature extraction between activities is challenging. In contrast, Figure \ref{T-SNE result} (e), (f), (g) and (h), which showcase the pre-trained model's results, display distinct separation or clusters for each category, indicating improved feature extraction. This capability contributes to enhanced classification performance.

\begin{figure}[htpb]
	\centering
	\subfloat[]{\includegraphics[width=0.45\textwidth]{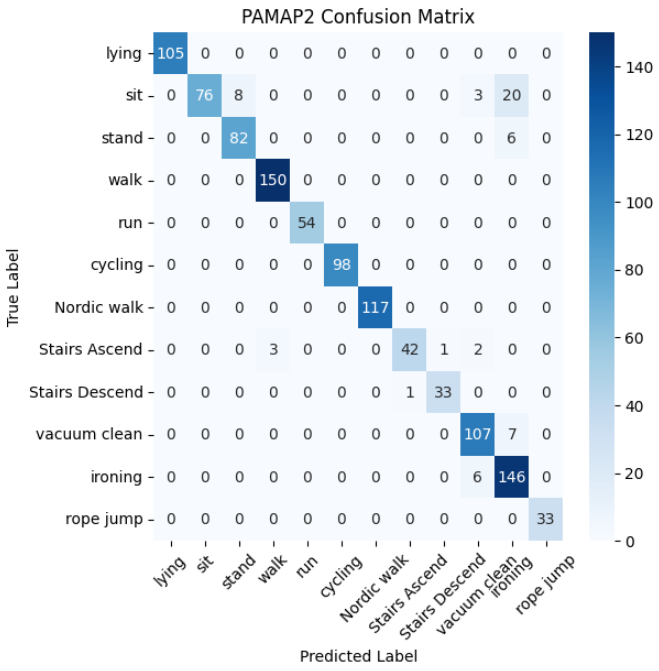}%
		\label{pamap2_mthar}}
	
	\subfloat[]{\includegraphics[width=0.45\textwidth]{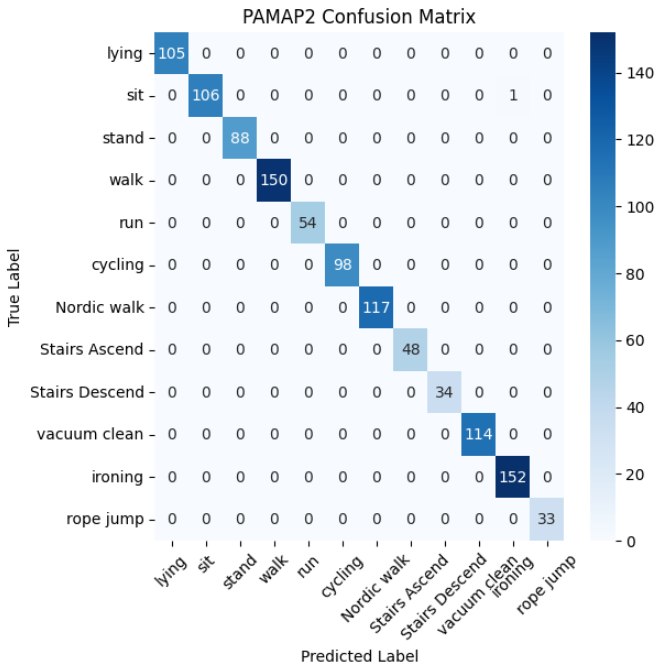}%
		\label{pamap2_harmamba}}
	\caption{The confusion matrices on the PAMAP2 dataset between MTHAR\cite{duan2023multi} and HARMamba. (a) MTHAR, (b) HARMamba}
	\label{confusion_pamap2}
\end{figure}

The figure \ref{confusion_pamap2} illustrates the confusion matrices for HARMamba and MTHAR on the PAMAP2 dataset. It demonstrates that our proposed framework significantly enhances the performance for activities like 'sit', 'stand', 'cycling', and 'ironing', among others. These findings substantiate the superior recognition capabilities of our framework.

\begin{table}[htbp]
    \renewcommand\arraystretch{1.5}
    \centering
        \caption{\label{param and Flop} Performance comparison of HARMamba and Channel-Equalization-HAR models on four datasets}
    \begin{tabular}{c|c|c|c}
        \toprule
        \bf{Method} & \bf{param.(M)} & \bf{FLOPs(M)}  & \bf{F1} \\
        \midrule
        % 添加PAMAP2行，并居中
        \multicolumn{4}{c}{\bf{PAMAP2}} \\
        \toprule
        Channel-Equalization-HAR & 1.6800 & 488.69  & 0.9218 \\
        % Transformer & 3.3060 & 211.06 & 512& 512\\
        HARMamba  & \bf{0.4271} & \bf{30.60}  & \bf{0.9974} \\
        \bottomrule
        % 添加UCI行，并居中
        \multicolumn{4}{c}{\bf{UCI}} \\
        \toprule
         Channel-Equalization-HAR & 0.4300 & 43.82  & \bf{0.9712} \\
        % Transformer  & 2.5057 & 197.91 & 128 & 128\\
        HARMamba & \bf{0.3883} & \bf{11.07}  & 0.9701\\
        \bottomrule
        % 添加UNIMIB行，并居中
        \multicolumn{4}{c}{\bf{UNIMIB SHAR}} \\
        \toprule
         Channel-Equalization-HAR & 0.4700 & 25.86  & 0.7869 \\
        % Transformer & 2.4637  & 162.26 & 151 & 151\\
        HARMamba  & \bf{0.3522} & \bf{7.94}  & \bf{0.8823}\\
        \bottomrule
        % 添加WISDM行，并居中
        \multicolumn{4}{c}{\bf{WISDM}} \\
        \toprule
         Channel-Equalization-HAR & 0.4300 & 34.86  & 0.9918 \\
        % Transformer  & 2.4796 & 219.71  & 200 & 200\\
        HARMamba  & \bf{0.2201} & \bf{10.58}  & \bf{0.9920}\\
        \bottomrule
    \end{tabular}
\end{table}
\textit{2) Performance evaluation}

To assess the performance of the HARMamba model, we compared it with the Channel-Equalization-HAR model \cite{huang2023channel} on four benchmark datasets. The aim was to compare the number of model parameters (M) and the amount of calculation FLOPs (M) in the forward propagation process of the model. Channel-Equalization-HAR is a light-weight convolutional neural network. The comparison results between the two models are presented in Table \ref{param and Flop}. HARMamba achieves equivalent or higher recognition accuracy and F1 scores than the Channel-Equalization-HAR model on all four datasets, with lower model parameters and FLOPs. This suggests that our model has lower computational costs and is suitable for real-time activity recognition on mobile devices.

% The HARMamba block module was replaced with the Transformer architecture, and the performance of both models was evaluated on four datasets, including the number of model parameters and FLOPs. Table \ref{param and Flop} presents the comparison results between the two models. On four datasets, HARMamba has fewer parameters than the Transformer model, and it has higher FLOPs with the same patch size.

\begin{figure}[htpb]
	\centering
	\includegraphics[width=0.5\textwidth]{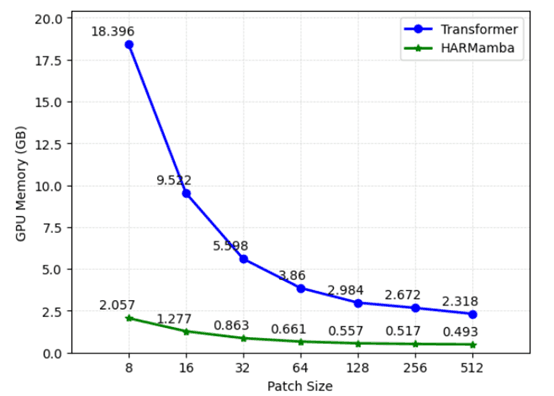}
	\caption{The memory efficiency results of Transformer and HARMamba models are compared under different patch sizes on the Pamap2 dataset.}
	\label{gpu figure}
\end{figure}

To evaluate the efficiency of HARMamba in activity recognition tasks, we benchmarked GPUs memory on the PAMAP2 dataset using different patch sizes. Figure \ref{gpu figure} shows that we replaced the HARMamba Block module with the standard Transformer module to test its memory resource consumption. Benefiting from the hardware-aware design of SSM, HARMamba has excellent linear scaling performance and consumes less computing resources than the Transformer block.
% To evaluate the efficiency of HARMamba in activity recognition tasks, we benchmarked GPU memory on the PAMAP2 dataset using different patch sizes. Figure \ref{gpu figure} shows that we replaced the HARMamba module with the Transformer module to use memory resources. Benefit from the hardware-aware design of SSM, HARMamba has excellent linear scaling performance and consumes fewer computing resources than the Transformer model.

\subsection{Ablation experiment}

\subsubsection{The impact of bidirectional SSM on classification results} Three strategies are mainly studied: (1) SSM: using a single Mamba block to process the sensor data representation sequence directly. (2) Bidirectional SSM: adding an additional SSM to each block to handle the sequence of sensor data representations in the backward direction. (3) Bidirectional SSM + Conv1D: adding a Conv1D before each SSM based on Bidirectional SSM.

% As shown in Table \ref{ssm impact}, a single SSM can achieve good recognition results, but it lacks understanding of the global sequence context modeling, and the best performance is achieved by adding backward SSM and \textcolor{red}{Conv1D}. The default strategy in the HARMamba block is to use bidirectional SSM+\textcolor{red}{Conv1D}.

HARMamba incorporates a bidirectional SSM, enabling the model to process sensor data sequences in both forward and backward directions. This design facilitates the simultaneous consideration of both preceding and following information at each position within the sensor sequence, thereby enhancing the model's ability to represent information accurately for activity recognition tasks. The bidirectional approach improves the model's comprehension of the context in the data, allowing it to more effectively capture temporal dependencies. The inclusion of past and future data points provides a more comprehensive representation of the activity being identified. As shown in Table \ref{ssm impact}, a single SSM can achieve good recognition results, but it lacks understanding of the global sequence context modeling, and the best performance is achieved by adding backward SSM and Conv1D. The default strategy in the HARMamba block is to use bidirectional SSM + Conv1D

\begin{table}[htbp]
	\renewcommand\arraystretch{1.5}
	\centering 
	\caption{\label{ssm impact}The impact of different strategies on the results on the UNIMIB SHAR dataset}
	\begin{tabular}{lcc}
		\toprule
		 Strategy & Accuracy & F1 score \\
		\midrule
            SSM &  77.30\% & 0.7723 \\
		  Bidirectional SSM &  85.57\% & 0.8569 \\
            Bidirectional SSM + Conv1D &  \bf{88.08\%} & \bf{0.8823} \\
		\bottomrule
	\end{tabular}
\end{table}

\subsubsection{Comparison of Results from Channel Independent and Channel Fusion Methods} To assess the channel independent strategy's influence on performance, we performed experiments comparing the activity recognition capabilities of HARMamba models employing channel fusion and channel independence. Utilizing PAMAP2, UCI, UNIMIB SHAR and WISDM datasets, with 9, 9, 3, and 3 sensor channels respectively, the results are compiled in Table \ref{channel strategy impact}. HARMamba processes each sensor channel independently, which isolates the impact of noise or invalid data from one channel from others. This independence allows the model to leverage valid data from other channels, thereby enhancing its resilience to sensor failures or inaccuracies. Table \ref{channel strategy impact} demonstrate that this strategy significantly enhances the F1 score for all datasets, indicating that not all channels contribute equally to the recognition task.

\begin{table}[htbp]
	\renewcommand\arraystretch{1.5}
	\centering 
	\caption{\label{channel strategy impact} F$_{1}$ score results of channel-independent and channel-fusion methods on four benchmark datasets}
	\begin{tabular}{lcccc}
		\toprule
		\bf{Methods} & PAMAP2 & UCI & UNIMIB &WISDM \\
		\midrule
    Channel Fusion& 0.9876 & 0.9493 & 0.8683 & 0.9634\\
		Channel Independent & \bf{0.9974} & \bf{0.9701} & \bf{0.8823} & \bf{0.9920} \\
		\bottomrule
	\end{tabular}
\end{table}

\subsubsection{Classification design} We ablated the classification design of HARMamba and used the four datasets of PAMAP2, UCI, UNIMIB SHAR and WISDM as benchmarks. We study the following classification strategies:
\begin{itemize} 
	\item No class token. we forego the addition of class tokens to the input sequence. Instead, we focus on extracting features from the final HARMamba block, applying mean pooling, and subsequently performing classification on the pooled output.
	\item End classification token. We concatenate the class token at the tail of the token sequence and perform classification.
\end{itemize}

% As demonstrated in Table \ref{Classification design}, experiments have shown that the classification token strategy, when added at the end of the sequence, can optimally utilise the recursive nature of SSM, resulting in the highest accuracy on four datasets.
The introduction of classification token enables the model to aggregate and integrate information from different patches following the processing of all input data. This aggregation facilitates a better understanding of the context of the entire sequence, thereby enhancing classification accuracy. The experimental results presented in Table \ref{Classification design} demonstrate that the strategy of employing classification token can significantly improve classification accuracy across multiple datasets. For instance, on the PAMAP2 dataset, the model utilizing classification token attained an accuracy of 99.91\%, whereas the model without classification token achieved only 97.70\%. This illustrates the pivotal role of classification token in enhancing model performance.

\begin{table}[htbp]
	\renewcommand\arraystretch{1.5}
	\centering 
	\caption{\label{Classification design} Accuracy results of different classification designs on four benchmark datasets}
	\begin{tabular}{lcccc}
		\toprule
		\bf{Classification strategy} & PAMAP2 & UCI & UNIMIB & WISDM \\
		\midrule
    No class token& 97.70\% & 94.53\% & 82.30\% & 95.09\%\\
  End class token & \bf{99.91\%} & \bf{97.65\%} & \bf{88.08\%} & \bf{99.11\%} \\
		\bottomrule
	\end{tabular}
\end{table}

\section{Conclusion and future work}
% We introduce HARMamba, a light-weight activity recognition model that employs the state-of-the-art Mamba architecture as its backbone. By implementing bidirectional state space modeling on sensor patch blocks, HARMamba outperforms attention-based and convolutional networks in terms of recognition accuracy while maintaining lower computational complexity and memory consumption. This demonstrates its potential as a favorable backbone for activity recognition tasks, with plans for future extensions. Specifically, we aim to apply Mamba in self-supervised learning to alleviate the need for active labeling, explore cross-human domain adaptation, and deploy the model on mobile devices for real-time activity recognition.
We propose the HARMamba architecture, an innovative state-space model-based sequence representation learning framework for wearable sensors. The HARMamba model employs sequence modeling to learn sensor data. By incorporating bidirectional state space modeling, HARMamba not only achieves data-driven global context modeling but also maintains modeling capabilities comparable to other models such as the Transformer, while significantly reducing computational complexity. Furthermore, owing to the hardware-aware optimization algorithm of the Mamba architecture, HARMamba significantly enhances inference speed and memory usage efficiency when processing long sensor sequences, underscoring its considerable potential to emerge as the next-generation HAR backbone network.

In future work, the bidirectional state space modeling technology of the HARMamba model provide new possibilities for self-supervised learning tasks. Examples include exploring self-supervised learning techniques to reduce the need for labeled data, enhancing cross-domain adaptability, and investigating applications of HARMamba in other areas beyond HAR, such as gesture recognition or environmental monitoring. HARMamba's superior performance in human activity recognition can be attributed to its innovative use of bidirectional SSM, independent channel processing, and patching. These features not only improve the accuracy and robustness of the model but also make it suitable for real-time applications. 

% \bibliographystyle{IEEEtran}

% \bibliography{IEEEabrv,references}
% \end{thebibliography}

% \vfill

\bibliographystyle{IEEEtran}

\end{document}